# Assessing Lower Limb Strength using Internet-of-Things (IoT) Enabled Chair

Authors: Chelsea Yeh[(i)], Hanna Kaitlin Kalaw Dy[(ii)], Phillip Schodinger[(ii)], Hudson Kaleb Dy[(iii)]

Affiliations: Yale University[(i)], Walnut High School[(ii)], California Institute of Technology[(iii)]

*Abstract*—**We propose an IoT enabled chair to dynamically and continuously monitor weight distribution of the seat in the goal of assessing lower limb strength in patients without the need for a physician to be present. This project describes the application of the technologies of Internet-of-Things (IoT) to assess the lower limb strength of individuals undergoing rehabilitation or therapy. Specifically, it seeks to measure and assess the progress of individuals by sensors attached to chairs and processing the data through Google GPU Tensorflow CoLab. Pressure sensors are attached to various locations on a chair, including but not limited to the seating area, backrest, hand rests, and legs. Sensor data from the individual performing both sit-to-stand transition and stand-to-sit transition provides a time series dataset regarding the pressure distribution and vibratory motion on the chair. The dataset and timing information is fed into a machine learning model to estimate the relative strength and weakness during various phases of the movement. They are compared with previously collected data to determine the efficacy and progress of the rehabilitation or therapy. This project created IoT-connected pressure and motion sensors. These were then used to capture the individual's weight distribution and motion on the chair and send the data to a central server. This data is then made available to be processed in Google GPU CoLab. The results of this research are displayed as preliminary data which shows that the chair differentiates when a person with a simulated leg injury (preference given to one leg over another) sits and stands. There are also variations in the attack rates in the graphs when different people sit and stands up. This provides promise that a future multi-disciplinary research project can collect and catalog data for training a multivariate time-series machine learning classification algorithm. Ultimately, this would allow continuous, in-home monitoring of at-risk populations like elderly adults and proactively prevent falls.** (*Abstract*)

## I. INTRODUCTION

During physical therapy, rehabilitation, or training, the measurement of lower body strength provides indications of the patient's physical therapy and rehabilitation progress. They also provide lower limb strength and endurance assessments of older adults. These tests include the 30-second chair stand test (30CST) and 5 times sit-to-stand test (5xSTS) [1,2]. Internet of Things (IoT) allows for real-time remote collection and interpretation of data [3] that can be stored on a central database and processed with cloud-based multivariate time-series classification machine learning processes [4].

In this paper, we present chairs equipped with pressure sensors will be able to capture the dynamic weight distribution and motion of the seat as the patients perform the test and provide time-series data regarding the area of performance improvement and strengthening. The collected data is sent via custom web services where the data is stored and aggregated. This data is then fed into and graphed by Google's GPU CoLab. The goal of this research is to eventually create a new tool to classify lower limb strength and endurance quantitatively and accurately by measuring the relative dynamic forces measured by the sensors on the chairs. In this way, the progress of the treatment or impairment can be assessed without using the current standard of requiring physicians or licensed physical therapists to diagnose patients.

## II. SIT-TO-STAND ASSESSMENT

The current standard of assessing functional lower extremity strength in older adults is the Sit-to-Stand Test (STS).[16] The two most common variants of the Sit-to-Stand Test are the Five Times Sit-to-Stand Test (5TSTS) and the 30 Second Sit-to-Stand Test (30CST). The 5TSTS measures how many seconds a patient needs to complete 5 Sit-to-Stand motions while the 30CST counts how many times they can stand up from a sitting position repeatedly over the course of 30 seconds. These assessments are currently conducted by professional physical therapists and/or physicians who will then assign them a score to classify the user.[17]

This project involved adding strain sensors and storing and processing time-series data to evaluate lower limb strength/mobility. We also automatically segment and pre-process the collected data to prepare for easier analysis of the results to potentially train a machine learning model in a future project.

The goal is to create an automated system that will improve upon the existing standard of using a "30 Second Sit-to-Stand Test" that must be performed by a licensed medical practitioner.

There are nuances in the time-series weight distribution data in the four corners of a chair based on a person's lower limb strength. These nuances would be too minute for humans to recognize, but it may be sufficient for a deep learning convolutional network machine learning algorithm to properly classify.

Our proposal is an automated system that can accurately record time-series data and be able to graph these data in Google CoLab cloud environments. Doing so would allow future multi-disciplinary research with licensed physical therapists or physicians to collect and identify training data of patients.

The hope is that the IoT chair we developed would lead to future multi-disciplinary research that eventually creates a new modality in quantitatively assessing patient lower-limb strength and mobility using an automated machine learning process instead of the current standard of using Sit-to-Stand Tests that requires medical professional intervention. The dense time series can be a valuable tool in the assessment and treatment of lower-body injuries, disease, and aging, and may have applicability in sports or fitness training.

## III. COMPONENTS

The IoT lower body assessment system consists of the following elements:

1) A chair suitable for performing sit-to-stand testing that can be equipped with sensors
2) Network of pressure sensors and motion sensors suitable for mounting to a chair;
3) Internet-capable processing devices and software co-located and connected to the chair sensors, which collect, interpret, and transmit the sensor data
4) A Linux Apache web server that hosts custom PHP web services. These web services are used by the IoT chair to store data to a MySQL Server. It also performs post-processing and feeds the data to Google's Tensorflow/Keras GPU Colab instances.
5) A MySQL server that stores the chair sensor data and processes them into the required format.
6) Google Colab is used to graph the result sets of each stand-to-sit and sit-to-stand sequence so that a user can analyse the data.

*A. Chair*

The chair requirement dictates that pressure sensors is mounted on different locations of the chair that will bear the full weight of the individual. To measure the dynamic weight distribution during the sit-to-stand transition, the accurate transmission of the forces to the sensors is required. We chose a chair with a flat hard surface without cushions. Sensors were mounted to be optimally distant from each other to provide the most independent measurement of the forces.

The sensor can be mounted in the following locations: 1) under the 4 legs of the chair, 2) between the legs of the chair and the frame, and 3) between the frame and the seat of the chair. Mounting the sensors under the legs of the chair can cause the chair to be unstable and unsafe for the user. For most chairs, including folding chairs, there is no way to disassemble the chair in such a way as to mount the sensor between the legs and the frame due to construction. A non-folding wooden chair was sourced from a DIY furniture store [6] that had a single hard seating surface that is screwed on the frame. This met all of the measurement requirements and provided a good mounting location for the sensor between the seat and frame.

*B. Sensors*

For this project, strain gauges were chosen to measure the dynamic forces during the sit-to-stand transition. Strain gauges are electrical resistive elements mounted to rigid metals pieces that deform (strain) under forces (stress). As the metal piece strains under the weight, the resistance changes and is measured electrically.

To measure human weight, strain gauges each with a maximum range of 50kg (110lb) were chosen. These gauges are commercially available and commonly used for bathroom scales [7].

The electrical resistance of the strain gauges is measured via a quarter bridge circuit that converts the change in resistance to a change in voltage. The voltage is measured digitally by an analog-to-digital converter (ADC). The ADC used was the HX711 integrated circuit produced by Avia Semiconductor [8]. Designed for human weight measurements, it has a relatively fine analog resolution of 24 bits, but a relatively slow speed of 10Hz or 80Hz.

The modules come pre-configured for 10Hz measurement. Initial tests were performed at 10 samples per second and the smaller quantity of data is adequate for measurements thus far. However, the modules have also been tested at 80Hz and work equally well at the higher sample rate.

The HX711 ADC module is designed so 1, 2, or 4 strain gauges can be connected to it, and the reading will be the sum of the readings of the strain gauges. For bathroom scales, four (4) strain gauges are connected to one (1) ADC to give a single weight measurement. For this application, multiple weight measurements on different locations of the chair are needed. The initial attempt was to connect one (1) strain gauge for each ADC. However, the measurement drifted over time due to changes in environmental conditions such as temperature. So, the investigation was modified to use 2 strain gauge per ADC configuration. Here the strain gauges are connected differentially, so the drift from one device is canceled out by the drift from another device. This effectively removed the drift problem.

Four strain gauges were mounted to the four corners of the chair. They are connected to 4 ADC modules. Four (4) additional strain gages are connected differentially to the ADC modules, do not bear any weight and are unused for the measurement, and are used solely for drift cancellation.

*C. Microcontroller*

The microcontroller and software running on it are responsible for the user interface, data collection, and data transmission. The sensors are connected via a compatible interface to the microcontroller.

The microcontroller is equipped with networking capabilities, preferably wireless, which will be used to transmit the collected data to a server. It is powered via AC power or batteries.

The microcontroller chosen was the ESP32 because it is powerful, low cost, and has integrated WiFi capabilities [9].

The HX711 ADC's were connected to the ESP32 with a simple serial interface. WiFi networks are ubiquitous and secured with WPA2 [10]. The IoT devices are connected over a 2.4GHz WiFi band because it is more secure and has a longer range compared to the 5.0GHz band.

Custom software was developed on the Arduino platform to collect the sensor data. Custom drivers were required to read each of the four sensors asynchronously. While each ADC operates nominally at 10Hz, they actually sample at slightly different rates. Additional sensors can be added with minimal effort. The first software plots the sensor readings for chair setup and testing. The second software provides the user interface to calibrate the sensors, collect the data, and for the operator to label or classify the data for supervised machine learning training.

Initially, a button press initiated a 5-second data collection sequence at 10Hz. The data is packaged and sent via an HTTP POST request to the cloud-based data collection server. While this worked to create our proof-of-concept, the user interface was error-prone and required significant manual post processing of the collected data. This will be a big issue if the chair is used in a large data collection institutional setting. We subsequently improved the software by automating the data collection and segmentation without any user interaction. This was done by caching the data collected and automating when to start and stop sending data to the server based on the pressure on sensors. Doing so not only made data collection much easier and less error-prone, but it also made the data more consistent and ready to be used for training a deep learning time-series multivariate machine learning model in the future.

*D. Data Collection Server*

The data collection server is comprised of two components:

*1)* A Linux Apache Webserver [11] that hosts the PHP web services. These web services are used by the IoT chair to store data in the database, with separate web services for Training and Test modes. Webservices are also used to pull data out of the SQL server by the machine learning processes, again with separate web services used for Training and Testing modes.

*2)* A MySQL [12] database that is used to store time-series data for both Training and Testing.

*E. Future Training of a Machine Learning Model*

The data stored is pre-processed and fully accessible from the cloud-based Google Colab where the graphs are generated from. This same Google Colab includes the Keras [13] module of Tensorflow [14] which could be used to train a multivariate time-series classification model after sufficient data is collected.

A key advantage of this system is the distributed architecture. The data collected is accessible using separate standard web services and Tensorflow can easily be replaced with PyTorch or Amazon ML.

## IV. METHODS AND RESULTS

The prototype chair that was built is fully functional and has only been used to collect and store unit testing data by the investigators and participants performing sit-to-stand actions. We have successfully recorded the unit test data and retrieved them systemically in Google's GPU CoLab where Tensorflow or other machine learning engines could be used to further perform machine learning training and classification in future multi-disciplinary research when more sufficient and relevant data are collected.

Figure 1 shows a photograph of the prototype chair.

Figure 2 shows a photograph of the prototype chair with the seat removed showing the strain gauge sensors, HX711 ADC's, and ESP32 microcontroller.

Figure 3 shows the system architecture and data flow diagram.

Figure 4 shows a close-up view of the prototype board and weight sensor.

The initial version of our application required us to manually call a webservice to start and stop a data collection sequence. We have since improved upon the software to automate data segmentation to make the data ready for training by a machine-learning model and have an improved user interface to simply data collection.

With the improved software, we recorded new stand-to-sit and sit-to-stand motions. Each user performed three stand-to-sit and sit-to-stand activity sequences on the chair and their results are shown at Figure 5-16. Looking at the graphs, we can clearly see that all the graphs for each person is clearly differentiated from one user to another (User 1: Figure 5, 6, 7; User 2: Figure 8, 9, 10; User 3: Figure 11, 12, 13; User 4: Figure 14, 15, 16).

Figure 5, 6, 7 shows graphs of the actual data collected for sample user 1 performing a stand-sit-stand motion sequence plotted from Google Colab. Figure 8, 9, 10 show similar activity with user 2, Figure 11, 12, 13 for user 3 and figure 14, 15, 16 for user 4.

For each graph in Figure 5 to 16, each of the four colored lines represent the strain (weight) experienced by each of the four sensors at each corner of the chair. It should be noted that while the users sat down and stayed seated for different periods of time in each of their three trial sessions, the graphs for each user show a clear pattern of unique bounces and attack rates of change. For example, all four users show a consistent resting weight preference towards a side of the chair that is unique to that user. User 1 exhibit a smaller exaggerated pressure on the red and green line (front sensors) when first sitting and a larger exaggerated pressure right before rapidly releasing pressure (standing up) in the same red and green lines.

User 2 similarly shows more weight at the front of the chair but displays a very even pressure as soon as they are seated and also with no noticeable extra depression before standing up.

User 3 on the other hand, shows that there is a general uptick before releasing pressure (standing up), which could point to a slight rocking motion because of a need for additional leverage to stand up. It should be noted that the blue line also shows extra depressions as soon as someone sits, then relaxes until they are about to stand up again. These are results that should be interpreted by medical professionals to properly interpret the graphs.

Unlike the first 3 users, User 4 shows significant bounces and variability from each session with large bounces before performing a sit-to-stand action.

The data collected shows the viability of using IoT chair to evaluate patients when coupled with machine learning. But medical professionals and anatomy experts must be engaged to properly interpret and classify different scenarios so the machine learning algorithm can be properly trained. Full consent has been received by the investigators/human participants.

## V. DISCUSSION

All software has been developed with the completed user interface to facilitate data collection of the Training and Testing data. Data can be collected at both 10Hz and 80Hz sampling rate, but initial unit testing does not show any appreciable differences in the plotted graphs by the higher sampling rates. Therefore, all unit tests have been performed at 10Hz to reduce the amount of data collected and speed up plotting performance.

The actual unit testing performed and shown on Figure 5 to 16 shows actual data collected at 10Hz from 4 different users. Multivariate time-series classification machine learning has not been performed as we do not have any data from actual patients with known lower-limb strength classified by medical professionals. The purpose of the different data collection shown as Figure 5 to Figure 16 is to validate that there is an appreciable difference in the displayed plots of the actual stand-to-sit and sit-to-stand data sequences from one user to another.

This validation will be performed visually by looking at the graphs generated for this project.

In the future, this validation will ideally be performed via machine learning such that the IOT chair will directly classify a user for different lower limb deficiencies so no graphs will need to be generated. This of course, will require enough data collected and those data properly classified by medical professionals to facilitate training of the machine learning classification model so the system can automatically evaluate an individual's lower limb strength instead of simply showing a graph like it does now.

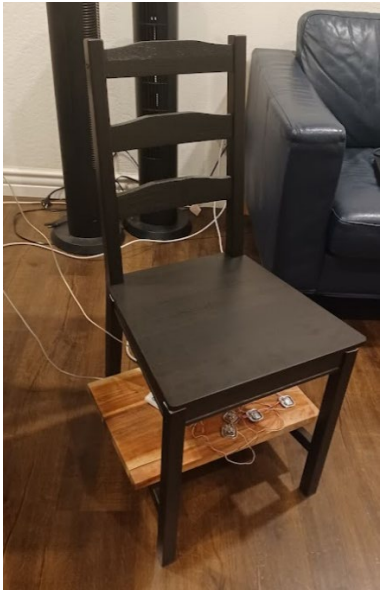

Fig 1. Picture of the Prototype.

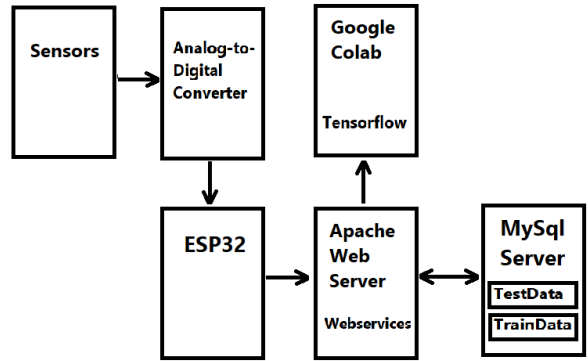

Fig 3. System Architecture and Data Flow Diagram

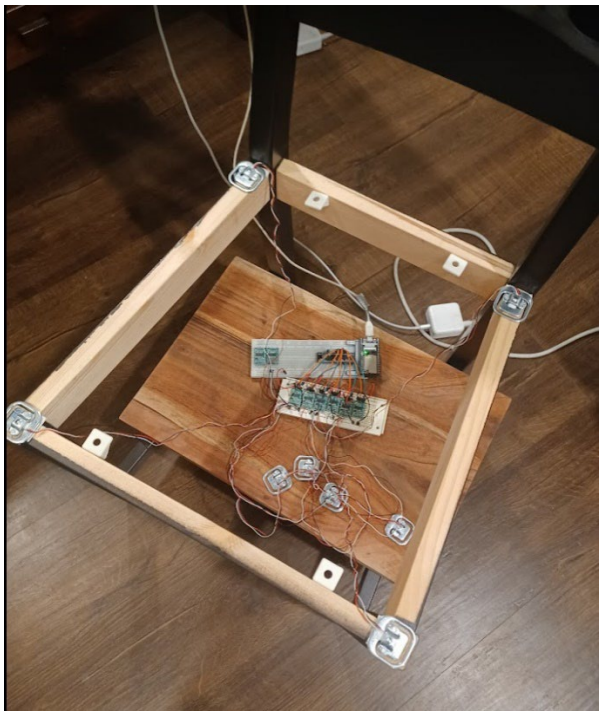

Fig 2. Prototype with seats removed showing sensors, ADC, and ESP32 module.

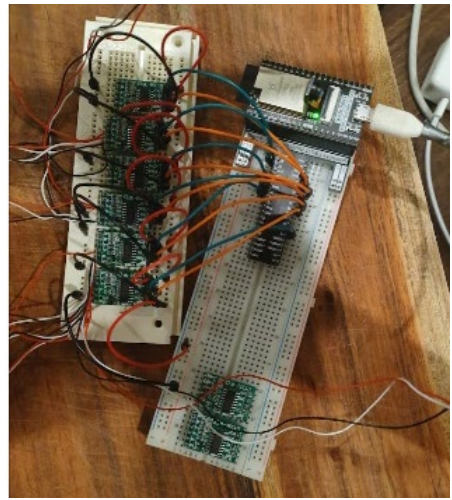

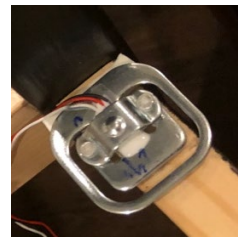

Fig 4. Prototype board (top) and weight sensor (bottom).

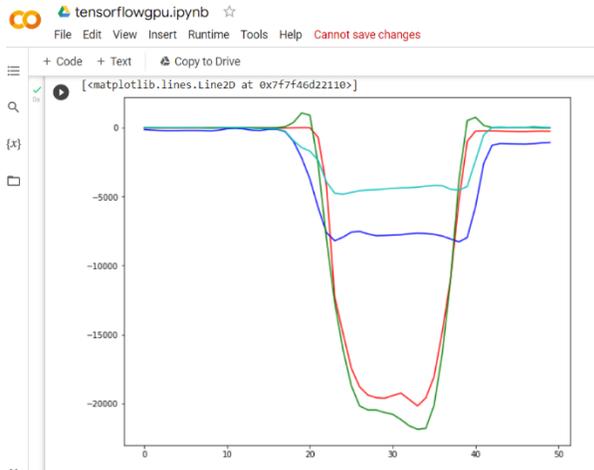

Fig 5. Sample Graph Of A 30 Second 10Hz Time-Series Dataset Plotting a Sit-To-Stand Motion with User 1, Iteration 1

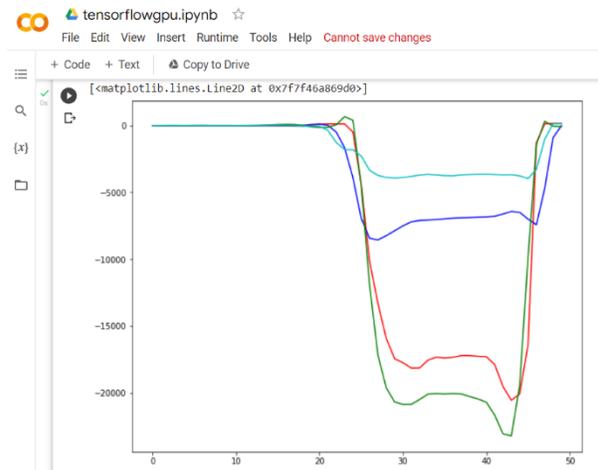

Fig 7. Sample Graph Of A 30 Second 10Hz Time-Series Dataset Plotting a Sit-To-Stand Motion with User 1, Iteration 3

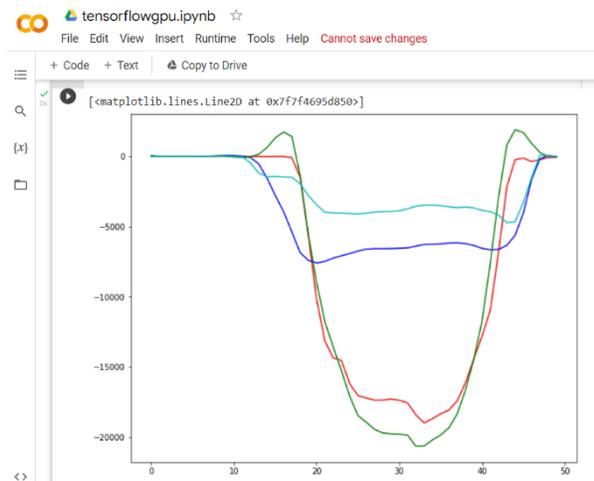

Fig 6. Sample Graph Of A 30 Second 10Hz Time-Series Dataset Plotting a Sit-To-Stand Motion with User 1, Iteration 2

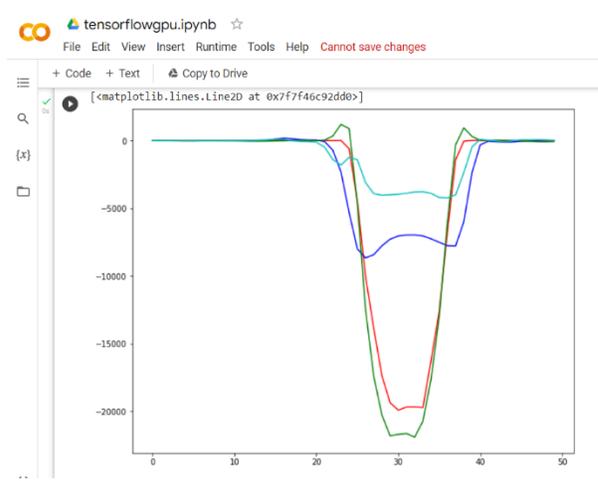

Fig 8. Sample Graph Of A 30 Second 10Hz Time-Series Dataset Plotting a Sit-To-Stand Motion with User 2, Iteration 1

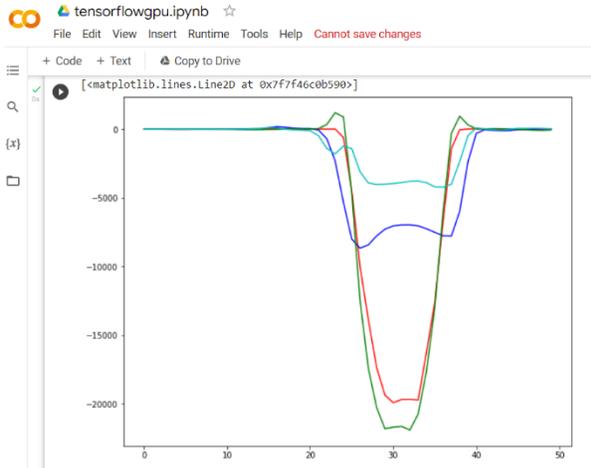

Fig 9. Sample Graph Of A 30 Second 10Hz Time-Series Dataset Plotting a Sit-To-Stand Motion with User 2, Iteration 2

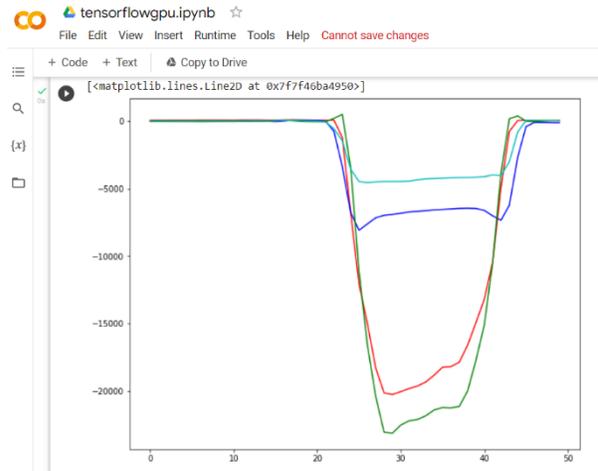

Fig 11. Sample Graph Of A 30 Second 10Hz Time-Series Dataset Plotting a Sit-To-Stand Motion with User 3, Iteration 1

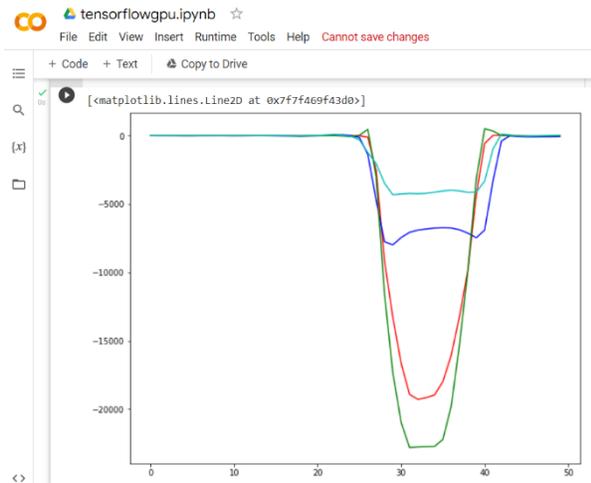

Fig 10. Sample Graph Of A 30 Second 10Hz Time-Series Dataset Plotting a Sit-To-Stand Motion with User 2, Iteration 3

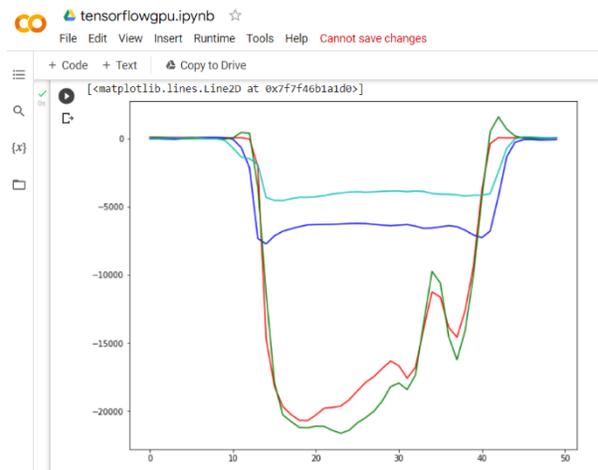

Fig 12 Sample Graph Of A 30 Second 10Hz Time- Fig 8. Sample Graph Of A 30 Second 10Hz Time-Series Dataset Plotting a Sit-To-Stand Motion with User 3, Iteration 2

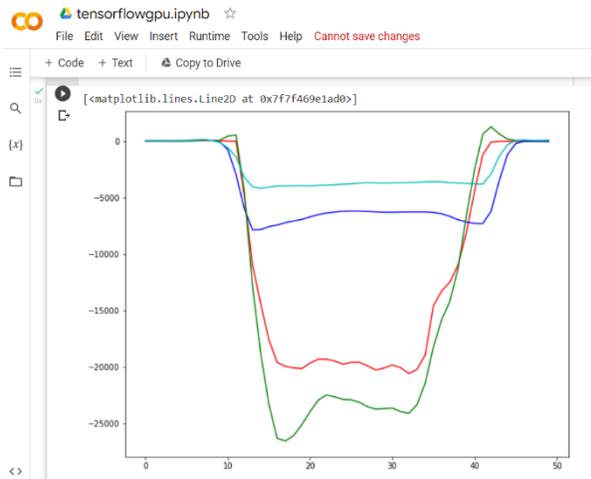

Fig 13 Sample Graph Of A 30 Second 10Hz Time- Fig 8. Sample Graph Of A 30 Second 10Hz Time-Series Dataset Plotting a Sit-To-Stand Motion with User 3, Iteration 3

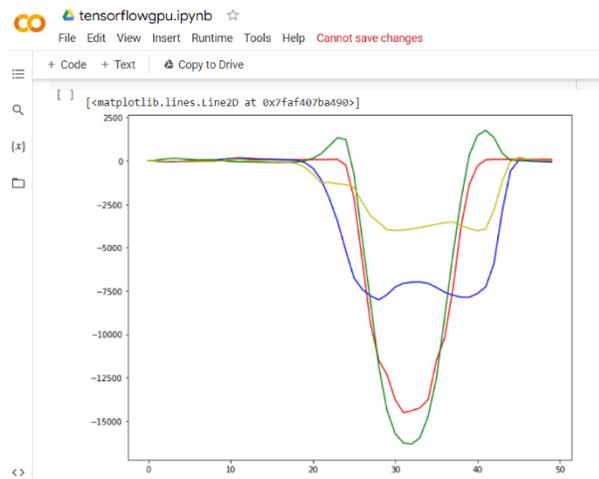

Fig 15 Sample Graph Of A 30 Second 10Hz Time- Fig 8. Sample Graph Of A 30 Second 10Hz Time-Series Dataset Plotting a Sit-To-Stand Motion with User 4, Iteration 2

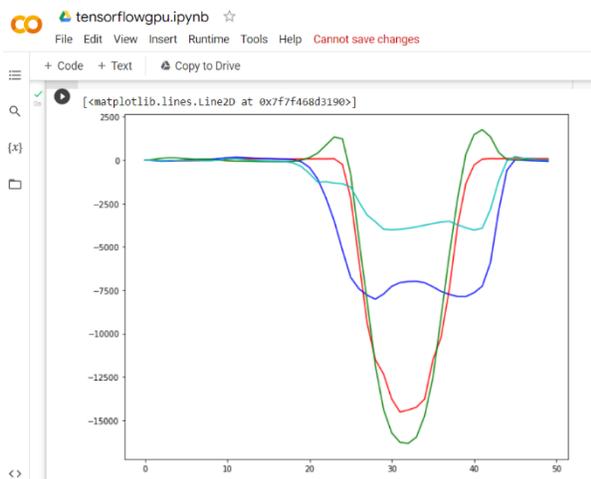

Fig 14 Sample Graph Of A 30 Second 10Hz Time- Fig 8. Sample Graph Of A 30 Second 10Hz Time-Series Dataset Plotting a Sit-To-Stand Motion with User 4, Iteration 1

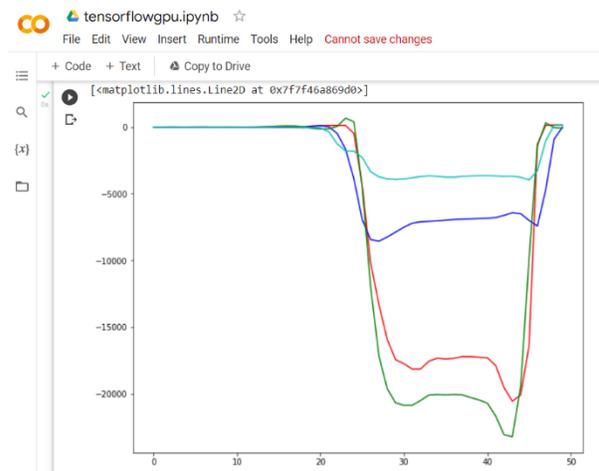

Fig 16 Sample Graph Of A 30 Second 10Hz Time- Fig 8. Sample Graph Of A 30 Second 10Hz Time-Series Dataset Plotting a Sit-To-Stand Motion with User 4, Iteration 3

## VI. Conclusion and Future Work

Our goal is to continue the project by partnering with medical professionals to collect samples from a large number of individuals of varying known strength/mobility as classified by medical professionals. These data can then be collected and made available to properly train and optimize the TensorFlow model.

Future plans include reaching out to kinesiology departments of local universities and finding faculty members that are interested in working with us. Optimization of the multivariate time-series classification models and creation of a standardized quantifiable measurement of lower limb mobility and strength will follow successful training of the machine learning model.

Further work could also include the estimation of isolated muscles or muscle groups from additional isolated muscle strength data collected during testing and training the machine learning model. This way the specific muscle groups for additional intervention can be identified.